\pdfoutput=1

\documentclass[conference,letterpaper]{IEEEtran}

\usepackage[utf8]{inputenc} 
\usepackage[T1]{fontenc}
\usepackage{url}
\usepackage{ifthen}
\usepackage{cite}
\usepackage[cmex10]{amsmath} 

\usepackage{hyperref}       
\usepackage{url}            
\usepackage{booktabs}       
\usepackage{amsfonts}       
\usepackage{nicefrac}       
\usepackage{microtype}      
\usepackage{lipsum}
\usepackage[ruled,vlined,linesnumbered]{algorithm2e}
\usepackage{setspace}
\usepackage{amsmath}
\DeclareMathOperator*{\argmax}{argmax}

\usepackage{amsthm}

\usepackage{graphicx}
\usepackage{subcaption}
\usepackage{xcolor}
\usepackage{cite}

\newcommand{\deepsic}{$\mathsf{DeepSIC}$}
\newcommand{\method}{$\mathsf{GANSIC}$}

\allowdisplaybreaks

\newtheorem{proposition}{Proposition}

\DeclareMathOperator{\sign}{sign}
\let\oldnl\nl
\newcommand{\nonl}{\renewcommand{\nl}{\let\nl\oldnl}}

\begin{document}
\title{Interference Cancellation GAN Framework for Dynamic Channels} 


\author{%
}

\author{%
	\IEEEauthorblockN{Hung T. Nguyen\IEEEauthorrefmark{1},
		Steven Bottone\IEEEauthorrefmark{2},
		Kwang Taik Kim\IEEEauthorrefmark{3},
		Mung Chiang\IEEEauthorrefmark{3},
		and H. Vincent Poor\IEEEauthorrefmark{1}}
	\IEEEauthorblockA{\IEEEauthorrefmark{1}%
		Princeton University,
		\{hn4,poor\}@princeton.edu}
	\IEEEauthorblockA{\IEEEauthorrefmark{2}%
		Northrop Grumman Corporation,
		steven.bottone@ngc.com}
	\IEEEauthorblockA{\IEEEauthorrefmark{3}%
		Purdue University,
		\{kimkt,chiang\}@purdue.edu}
}

\maketitle

\begin{abstract}
	Symbol detection is a fundamental and challenging problem in modern communication systems, e.g., multiuser multiple-input multiple-output (MIMO) setting. Iterative Soft Interference Cancellation (SIC) is a state-of-the-art method for this task and recently motivated data-driven neural network models, e.g. \deepsic{}, that can deal with unknown non-linear channels. However, these neural network models require thorough time-consuming training of the networks before applying, and is thus not readily suitable for highly dynamic channels in practice. We introduce an online training framework that can swiftly adapt to any changes in the channel. Our proposed framework unifies the recent deep unfolding approaches with the emerging generative adversarial networks (GANs) to capture any changes in the channel and quickly adjust the networks to maintain the top performance of the model. We demonstrate that our framework significantly outperforms recent neural network models on highly dynamic channels and even surpasses those on the static channel in our experiments.
\end{abstract}


\section{Introduction}

Symbol detection is the task of recovering the transmitted symbols from the noisy received channel output, and, is therefore fundamental in any communication systems. Multiuser multiple-input multiple-output (MIMO) setting \cite{marzetta2010noncooperative}, that supports high communication throughput demands of modern applications, makes this detection task substantially more challenging due to the interference of simultaneous transmissions of multiple symbols from a number of transmitters to multiple antennas through the same channel. Extensive research efforts have been devoted with numerous successes in both improving the detection accuracy in more realistic environments and efficiency of the algorithms to reduce communication delay.

Interference cancellation \cite{andrews2005interference} is a family of model-based symbol detectors that rely on channel model to cancel out interference. Iterative soft interference cancellation (SIC)\cite{wang1999iterative,alexander1999iterative,choi2000iterative} is an effective member in this family using soft symbol estimates to mitigate the error propagation. However, this algorithm assumes a linear Gaussian channel, in which noise follows a Gaussian distribution and interference is additive, and parameters of the noise distribution and channel matrix are known in advance. In many scenarios \cite{shlezinger2019asymptotic,studer2016quantized,iofedov2015mimo,khalighi2014survey,shlezinger2018capacity}, these assumptions are no longer valid. Considerable recent works \cite{farsad2018neural,caciularu2018blind,ye2017power,samuel2019learning,khobahi2019deep,he2018model,shlezinger2020deep} focus on data-driven neural network approaches to remove the dependency on channel model assumptions. \deepsic{} \cite{shlezinger2020deep} is among the most recent and has been shown great potential in various channel models including both linear and non-linear. More importantly, \deepsic{} resembles the architecture of SIC algorithm, and, thus, provides strong theoretical ground.

Despite the success of \deepsic{} in addressing the channel model assumption, it is quite time-consuming to train the entire network, and thus renders unsuitable for highly dynamic channels. We propose to combine \deepsic{} with the emerging generative adversarial networks \cite{goodfellow2014generative,mirza2014conditional} to design an online training model that can capture any changes in the channel and swiftly adjust the network accordingly. This ability to adapt keeps the model up-to-date and maintains the model's peak performance consistently. The online adaptation is possible in our model thanks to the characteristic that our model only requires noisy received signals from the channel to train. That is in contrast to earlier works which usually need both received signals and the corresponding transmitted symbols to train, as a typical supervised learning paradigm. In our experiments, we demonstrate better detection performance of our model consistently in highly dynamic and even static channels using various channel models.

\section{Preliminaries}

\begin{figure*}[!t]
	\centering
	\includegraphics[width=0.7\linewidth]{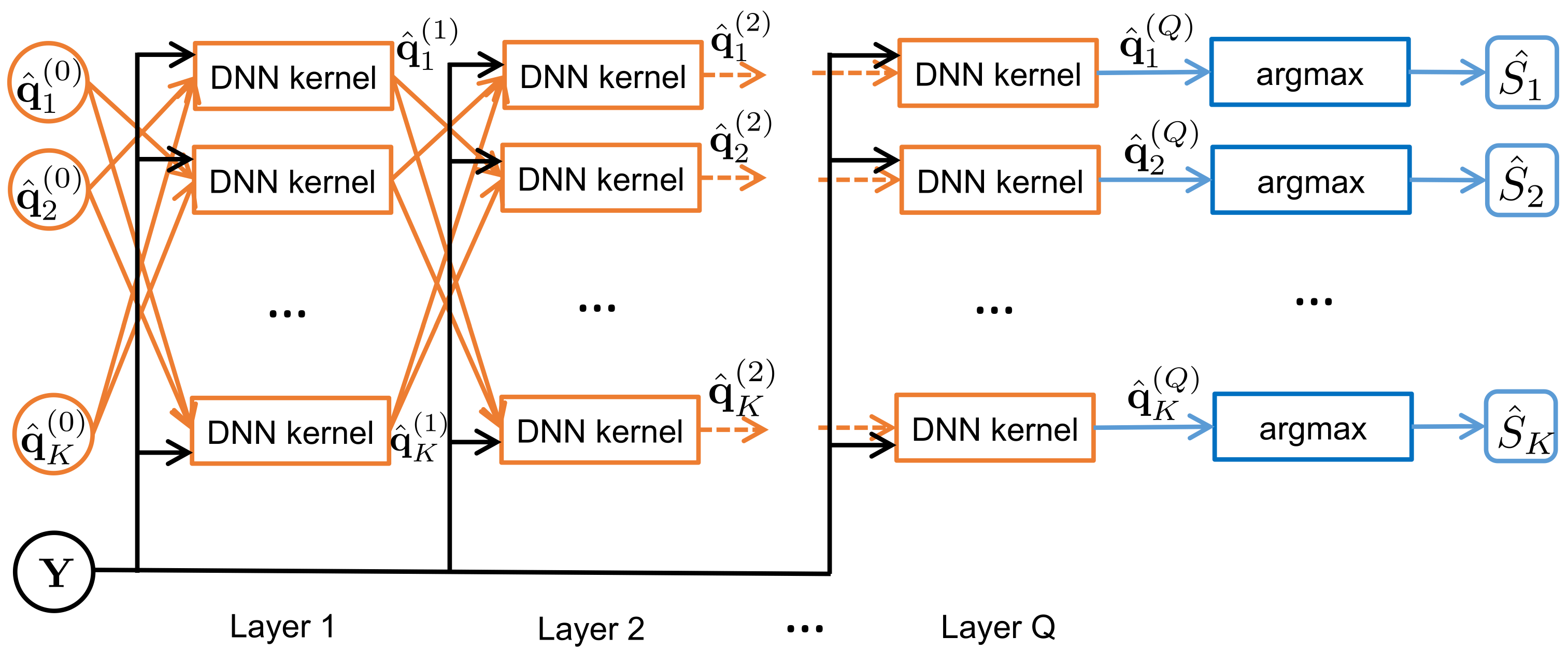}
	\caption{Illustration of \deepsic{} model.}
	\label{fig:deepsic}
\end{figure*}
\subsection{System model}
We consider an MIMO setting with $K$ transmitters and $R$ receive antennas over a memoryless stationary channel. Since the channel is memoryless and stationary, at any time instance, the transmitters send a real-valued vector $\mathbf{S} = [S_1, S_2, \dots, S_K]^T$, in which each element is randomly drawn from the constellation set $\mathcal{S}$ of information symbols. The receive antennas receives a real-valued channel output vector $\mathbf{Y} = [Y_1, Y_2, \dots, Y_R]^T$. Note that this system model also covers complex-valued channels as a vector of complex numbers can be represented by a higher-dimensional real-valued vector. The symbol detection task aims at recovering the vector of transmitted symbols $\mathbf{S}$ provided the received channel output $\mathbf{Y}$.

There are various channel models of how the channel produces its output $\mathbf{Y}$ from the transmitted vector $\mathbf{S}$. One of the common model is the linear Gaussian channel which consists of a linear transformation of $\mathbf{S}$ by a channel matrix $\mathsf{H}$ and an Additive White Gaussian Noise (AWGN) $\mathbf{W}$, i.e.,
\begin{align}
	\label{eq:awgn}
	\mathbf{Y} = \mathsf{H}\mathbf{S} + \mathbf{W}.
\end{align}
Many other non-linear channel models have also been considered: quantized Gaussian channel \cite{shlezinger2019asymptotic} modeling the low-resolution quantizers in wireless communication,
\begin{align}
	\label{eq:quant}
	\mathbf{Y} = q(\mathsf{H}\mathbf{S} + \mathbf{W}), \text{where } q(y) = \begin{cases}
	\sign (y), & |y| < 2 \\
	3 \cdot \sign (y), & |y| > 2
	\end{cases}
\end{align}
Poisson channel that model optical communications,
\begin{align}
	\label{eq:poisson}
	\Pr[Y_i | \mathbf{S}] = \mathbb{P}\left( \frac{1}{\sqrt{\sigma_w^2}} (\mathsf{H}\mathbf{S})_i + 1\right),
\end{align}
where $\mathbb{P}(\lambda)$ represents the Poisson distribution with parameter $\lambda > 0$ and $\sigma_w^2$ is the variance.

\subsection{Iterative SIC and \deepsic{} model}

Iterative SIC algorithm \cite{choi2000iterative} assumes a linear Gaussian channel with known channel matrix $\mathsf{H}$ and Gaussian noise distribution $\mathbf{W}$. It consists of $Q$ iterations and iteration $q$ produces $K$ distributions $\mathbf{q}_k^{(q)}$ for $K$ transmitters. Distribution $\mathbf{q}_k^{(q)}$ estimates the conditional probability of the transmitted symbol from transmitter $k$ given the channel output $\mathbf{Y}$. Through multiple iterations, the conditional probability estimates are gradually refined taking into account received signals $\mathbf{Y}$ and estimations $\mathbf{q}_k^{(q-1)}$ from the previous iteration and channel model assumptions. The detailed computations are in \cite{choi2000iterative}.

The \deepsic{} model is motivated by SIC algorithm and replaces each model-based update of conditional probability estimate $\mathbf{q}_k^{(q)}$ by a parameterized classification deep neural network (DNN) kernel. Thus, the entire model composes of $Q$ layers and each layer has $K$ of such DNN components. Each DNN kernel outputs probability estimate $\hat{\mathbf{q}}_k^{(q)}$ for transmitted symbol $S_k$ and takes in the estimates $\hat{\mathbf{q}}_j^{(q-1)}, j \neq k$ from the previous layer and the received signal $\mathbf{Y}$ as inputs. An illustration of \deepsic{} model is given in Figure~\ref{fig:deepsic}. To estimate the transmitted symbol $\hat S_k$ from output distribution $\hat{\mathbf{q}}_k^{(Q)}$ of the last layer, it returns the symbol with maximum estimated probability $\hat S_k = \argmax_{S \in \mathcal{S}} \hat{\mathbf{q}}_k^{(Q)} [S]$. The advantages of \deepsic{} compared to the original SIC algorithm arise from the fact that DNN kernel is agnostic from any channel model and learns the relationship between inputs and outputs from a set of training data. Thus, for any underlying channel, \deepsic{} only requires a training set of transmitted symbol and received signal pairs to train the network. This approach shares the same structure as iterative SIC, and hence offers a strong intuition. It was shown to perform comparably to iterative SIC on linear channel and substantially better on non-linear channels \cite{shlezinger2020deep}.

One of the problems with \deepsic{} is the intensive computation requirement. The network model involves many DNN kernels which all need to be properly trained before applying the network for symbol detection. Each training round needs a dataset of transmitted and noisy signal pairs collected before hand. In fast changing environments, \deepsic{} needs repetitive retraining causing low efficiency and serious communication delay. Therefore, this model is not readily suitable for highly dynamic channels in practice. In the next section, we introduce our online training model that does not require repetitive separate training and can quickly adapt to any changes in the channel to provide consistent highest performance of the detection network.

\section{Proposed model}

Our model combines deep unfolding approach in \deepsic{} with the emerging generative adversarial networks (GAN) \cite{goodfellow2014generative} which have revolutionized computer vision and many other fields in machine learning \cite{goodfellow2016deep}. Our model only asks for noisy received signals which are readily available and will swiftly adjust the network to keep it constantly up-to-date. This is in contrast with \deepsic{} that requires both received signals and their corresponding transmitted ones for training.

\begin{figure*}[!t]
	\centering
	\includegraphics[width=0.9\linewidth]{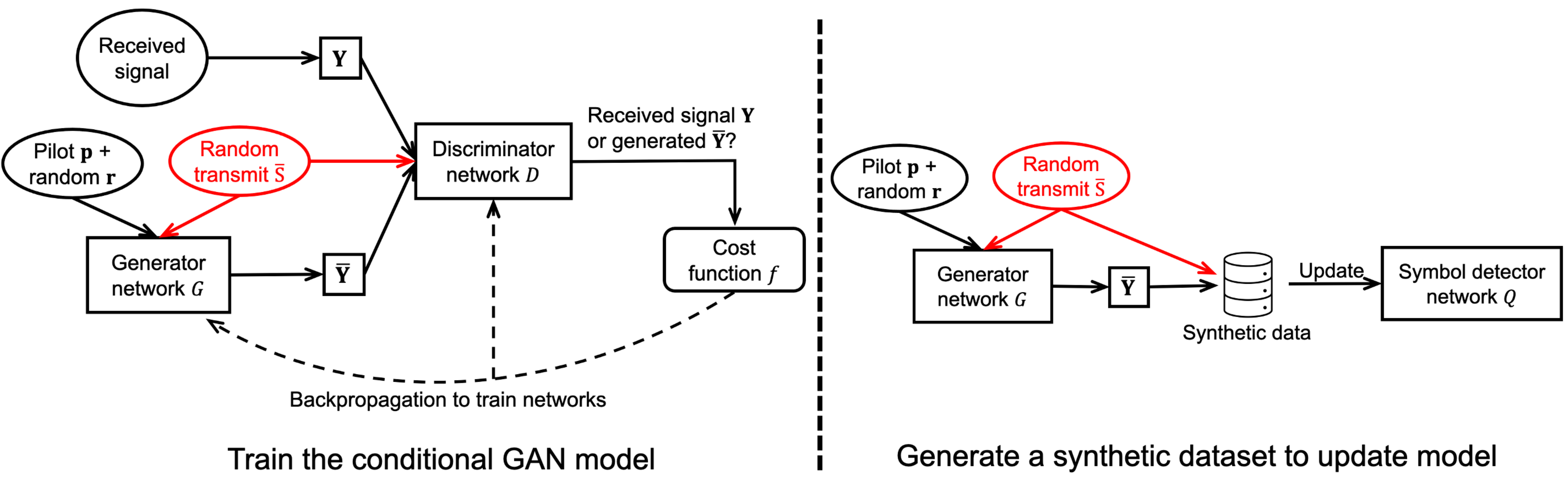}
	\caption{Illustration of the initial model.}
	\label{fig:initial}
\end{figure*}
\subsection{Initial model}

Our initial proposed model is depicted in Figure~\ref{fig:initial} that has the key component of a conditional GAN model aiming at generating synthetic noisy signals $\bar{\mathbf{Y}}$ resembling ones from current channel state. This conditional GAN uses pilot signals to capture the channel state information and randomly drawn transmit signals as the conditional variable feeding both the generator and discriminator networks. The generator $G$ uses the given pilot signal $\mathbf{p}$, random transmit symbols $\bar{\mathbf{S}}$ from the constellation set $\mathcal{S}$ along with a random sequence of real numbers $\mathbf{r}$ to generate a synthetic noisy signal that resembles the actual received signal from the random transmit symbols over the channel. The random sequence $\mathbf{r}$ diversifies synthetic outputs for the same random transmit symbols. 

The discriminator network $D$ gathers both real received signals from the channel and generated ones by the generator network along with the corresponding random transmit and tries to distinguish the real from the generated. The cost function $f$ evaluates how well the discriminator performs this classification task. Thus, intuitively, the generator $G$ and discriminator $D$ have competing objectives: generator $G$ strives to mimic the channel and generates outputs similar to actual received signals, while the discriminator $D$ wishes to differentiate real received signals from generated ones. In fact, the generator and discriminator participate in a zero-sum game and improve together as a better discriminator gives rise to a better generator in order to compete. 

A distinguished characteristic of our model is that it only needs noisy received signals from the channel to train which is the same input for symbol detection. Hence, the model, including both GAN components and symbol detector network, can be trained online during communications. Moreover, the symbol detector network $Q$ can be any trainable model, e.g., deep learning models, \deepsic{} \cite{shlezinger2020deep}, $\mathsf{DetNet}$ \cite{samuel2019learning}. In our experiments, we will use the most recent deep learning approach \deepsic{} for symbol detection.

\subsection{Properties and online training procedure}

First the cost function $f$ in training the conditional GAN model measures how well the discriminator classifies real received signals from generated ones. This function usually has the following form
\begin{align}
	f(G,D) = \mathbb{E}_{\mathbf{Y} \sim \mathbb{P}_C} [\log D(\mathbf{Y})] + \mathbb{E}_{\bar{\mathbf{Y}} \sim \mathbb{P}_G} [\log(1-D(\bar{\mathbf{Y}}))], \nonumber
\end{align}
where $\mathbb{P}_C$ denotes the distribution of received signals from the channel and $\mathbb{P}_G$ denotes the distribution of generated signals from generator $G$. Since the discriminator wants to maximize this function while the generator aims at minimizing it, we have the following optimization problem:
\begin{align}
	\label{eq:objective}
	\min_{G \in \mathcal{G}} \max_{D \in \mathcal{D}} f(G,D),
\end{align}
where $\mathcal{G}$ and $\mathcal{D}$ represent the spaces of generators and discriminators of interest. If we view the generator $G$ as a probability density function $\mathbb{P}_G$ and consider the space of all density functions, the following result follows from \cite{goodfellow2014generative}.
\begin{proposition}
	The optimal value of the optimization problem in \ref{eq:objective} is achieved if and only if $\mathbb{P}_G = \mathbb{P}_C$, or the distribution of synthetic signals matches the distribution of received signals from the channel.
\end{proposition}
Thus, if the GAN model is trained properly, the generator will be able to generate synthetic signals that resemble well ones from the channel. 

To train the GAN model, we alternatively train the generator and discriminator using stochastic gradient method. Particularly, a batch $\mathcal{B}_C$ of $m$ real received signal sequences from the channel and another batch $\mathcal{B}_G$ of $m$ generated ones by the generator $G$ are collected to calculate a estimate of loss function $f$ and its gradient, and then update the discriminator.
\begin{align}
	\label{eq:d_loss}
	f_D(G,D) = \frac{1}{m} \sum_{\mathbf{Y} \in \mathcal{B}_C} [\log D(\mathbf{Y})] + \frac{1}{m} \sum_{\hat{\mathbf{Y}} \in \mathcal{B}_G} [\log(1-D(\bar{\mathbf{Y}}))].
\end{align}
For the generator, only a batch $\mathcal{B}_G$ of $m$ generated signals are drawn to update the generator using the following loss.
\begin{align}
	\label{eq:g_loss}
	f_G(G,D) = \frac{1}{m} \sum_{\hat{\mathbf{Y}} \in \mathcal{B}_G} [\log(1-D(\bar{\mathbf{Y}}))].
\end{align}
Note that the gradient of $f_G(G,D)$ is back-propagated through both discriminator and generator networks when updating $G$.

The symbol detector network $Q$ is updated by drawing $n$ synthetic signals $\hat{\mathbf{Y}}$ along with the corresponding random transmit $\hat{\mathbf{S}}$ to form a batch of training pairs $(\hat{\mathbf{Y}},\hat{\mathbf{S}})$. When \deepsic{} is used for symbol detection, gradient descent is applied on sum-cross entropy loss and updates the network.

The overall online training procedures of GAN and updating symbol detector network $Q$ are described in Algorithm~\ref{alg:online_train}.
\begin{algorithm}[h]
	\caption{Online training GAN and updating symbol detector network $Q$.}
	\label{alg:online_train}
	
	\DontPrintSemicolon
	
	\setstretch{1.2}
	\SetKwInOut{Input}{Input}\SetKwInOut{Output}{output}
	\While{the channel is up}{
		Run \textsf{TrainGAN} and \textsf{UpdateDetector} in parallel.
	}
	\SetKwFunction{FGAN}{\textsf{TrainGAN}}
	\SetKwFunction{FDET}{\textsf{UpdateDetector}}
	\SetKwProg{Fn}{Procedure}{:}{}
	\Fn{\FGAN{}}{
		Gather $m$ received signal sequences from the channel.\\
		Generate $m$ synthetic signals $\bar{\mathbf{Y}}$ by generator $G$.\\
		Update the discriminator ascending its stochastic gradient $\nabla f_D(G,D)$ of Eq.~\ref{eq:d_loss}.\\
		Generate $m$ synthetic signals $\bar{\mathbf{Y}}$ by generator $G$.\\
		Update the generator descending its stochastic gradient $\nabla f_G(G,D)$ of Eq.~\ref{eq:g_loss}.\\
	}
	\SetKwProg{Pn}{Procedure}{:}{\KwRet}
	\Pn{\FDET{}}{
		Generate $n$ synthetic signals $\bar{\mathbf{Y}}$ by generator $G$ to form a training dataset $(\bar{\mathbf{Y}}, \bar{\mathbf{S}})$.\\
		Update the detector network $Q$ with the synthetic dataset $(\bar{\mathbf{Y}}, \bar{\mathbf{S}})$.\\
	}
\end{algorithm}

\begin{figure}[h]
	\centering
	\includegraphics[width=0.95\linewidth]{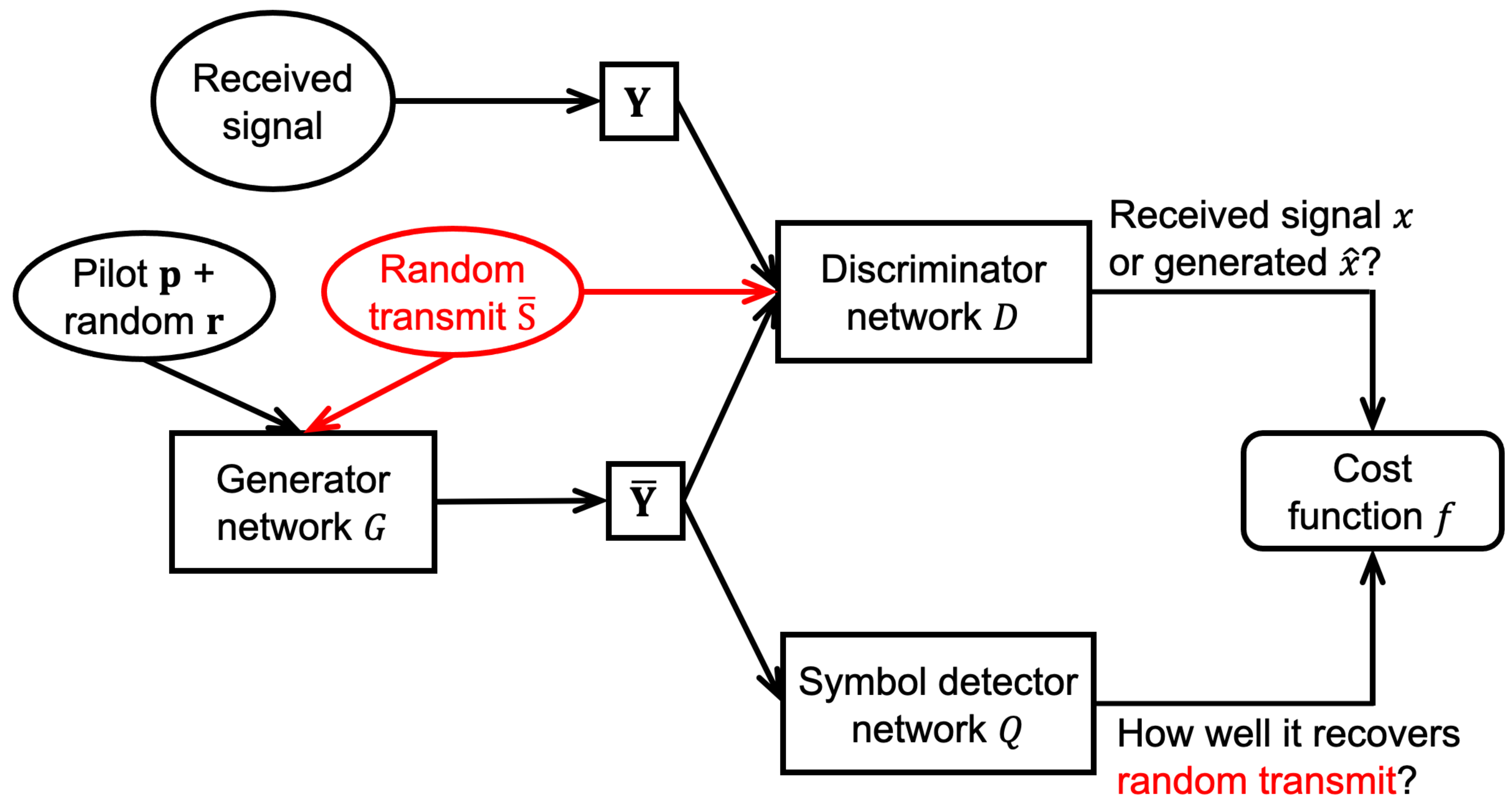}
	\caption{Efficiency-improved model.}
	\label{fig:improved}
\end{figure}
\begin{figure}[p]
	\vspace{-0.2in}
	\centering
	\begin{subfigure}{\linewidth}
		\includegraphics[width=0.95\linewidth]{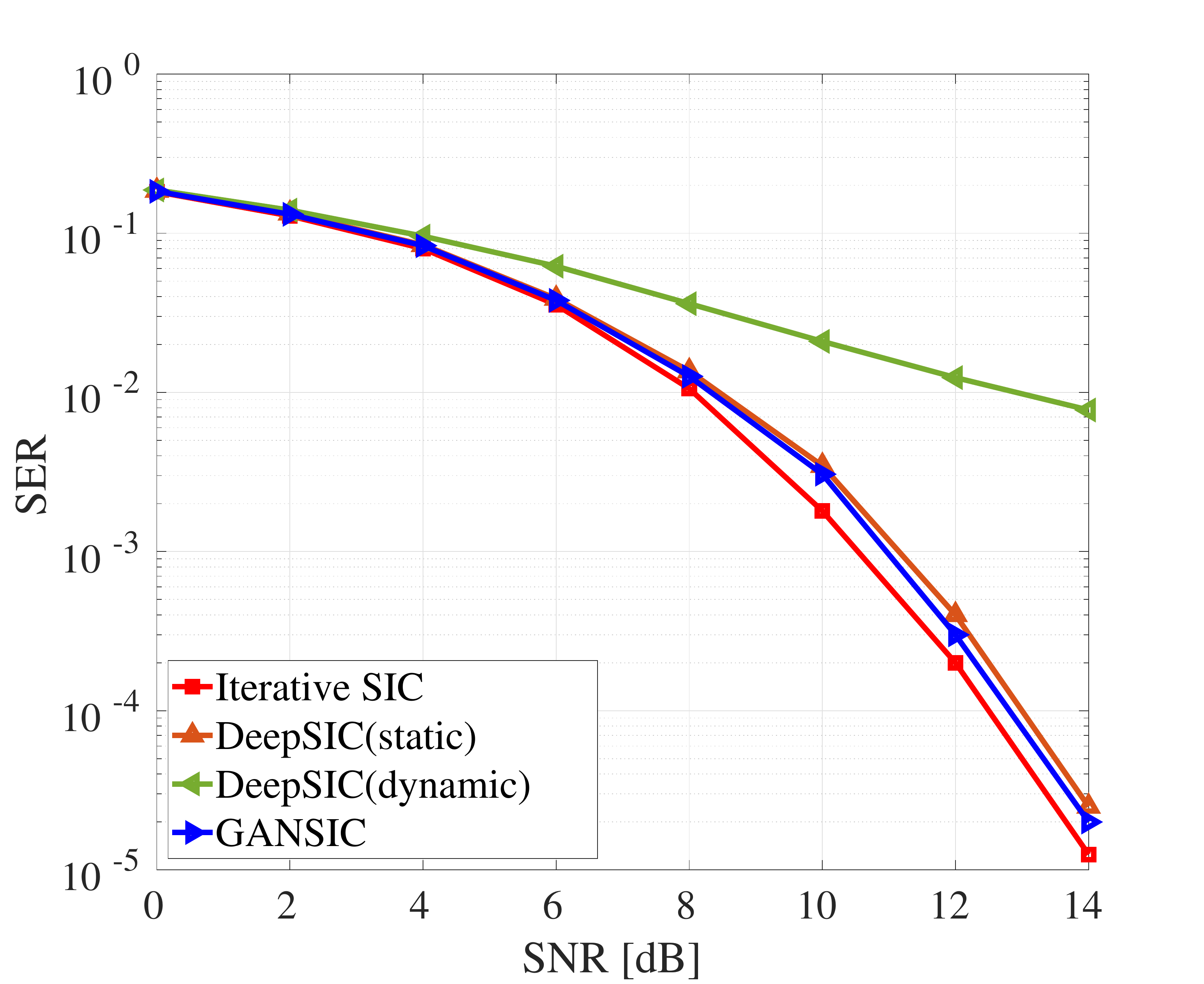}
		\caption{Linear Gaussian channel}
	\end{subfigure}
	\begin{subfigure}{\linewidth}
		\includegraphics[width=0.95\linewidth]{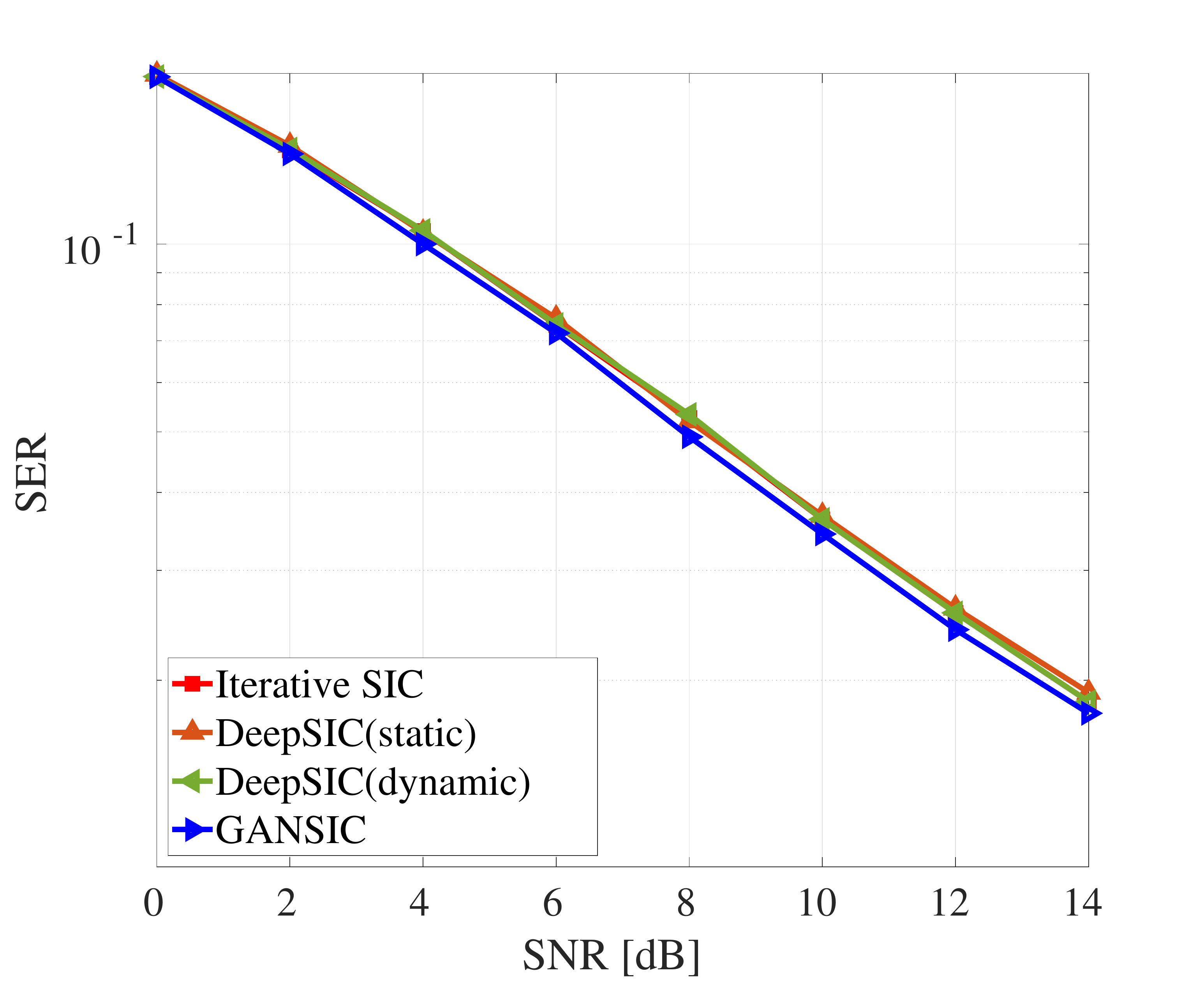}
		\caption{Quantized Gaussian channel}
	\end{subfigure}
	\begin{subfigure}{\linewidth}
		\includegraphics[width=0.95\linewidth]{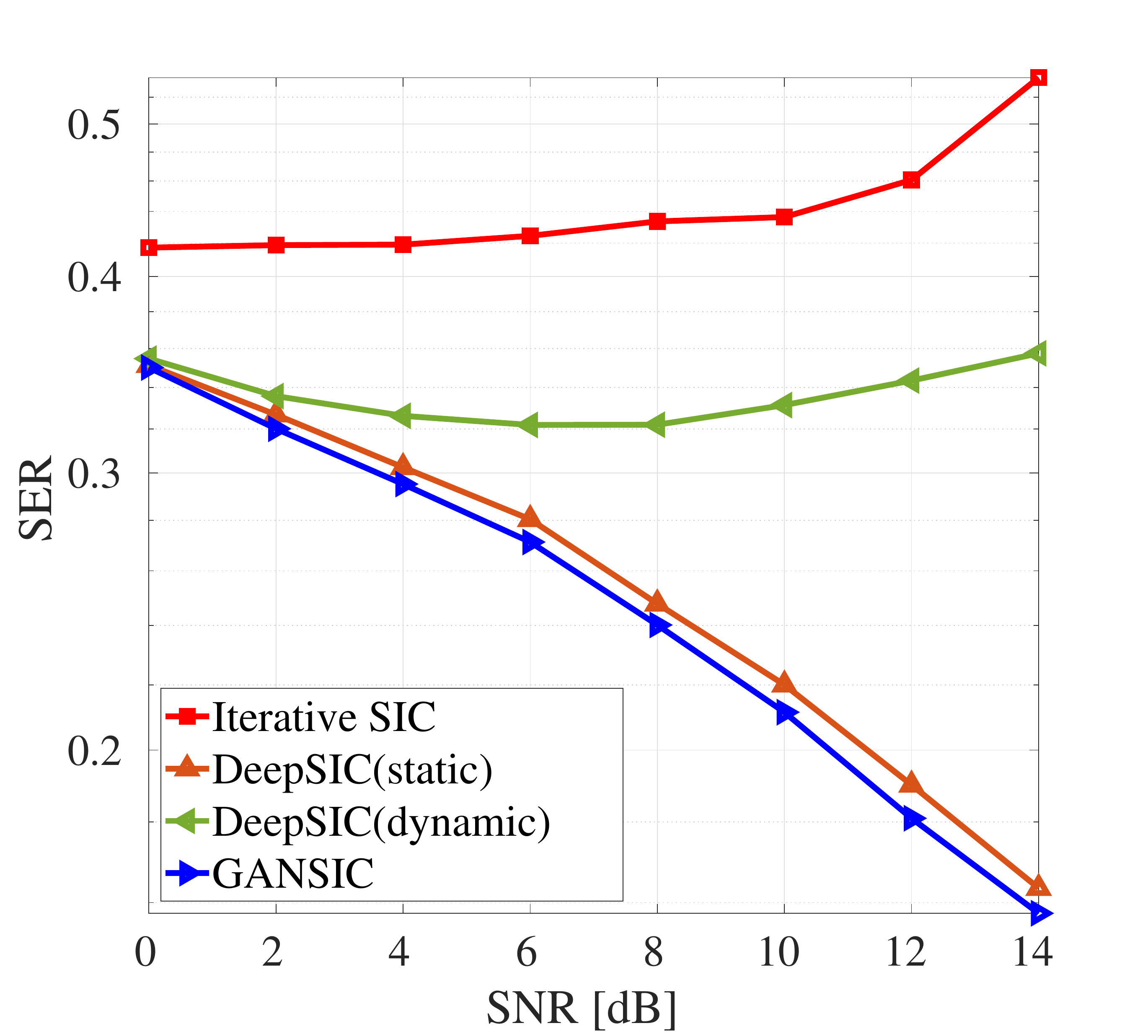}
		\caption{Poisson channel}
	\end{subfigure}
	\caption{Performance of different symbol detection methods on various linear and non-linear channels.}
	\label{fig:ser}
\end{figure}
\subsection{Efficiency-improved model}

Our initial model may still face an efficiency issue that it contains 2 separates updating of GAN and symbol detector networks. Here we propose to treat the symbol detector network as a component of GAN model. Our efficiency-improved architecture is described in Figure~\ref{fig:improved}. The cost function $f(G,D,Q)$ adds the original GAN cost with symbol detector cost and the batch $\mathcal{B}_G$ of generated signals is immediately used to update $Q$ as well as $G$. Thus, the training efficiency is improved by eliminating the separate $Q$ training procedure.

\section{Experiments}

In this section, we present our experimental results to demonstrate higher performance of our model on highly dynamic channels compared to existing data-driven neural networks and model-based approaches.

\subsection{Settings}
We consider \deepsic{} \cite{shlezinger2020deep} as symbol detector network in our framework and 2 layer neural networks with batch normalization and tanh activation functions for the generator and discriminator, and the middle layers have 512 neurons. For training, we use Adam \cite{kingma2014adam} with learning rate $\alpha = 0.0001$, decay rate $\beta = 0.5$, and batch size $m=n=64$.

We compare our model, namely \method{}, with the most recent deep neural network approach \deepsic{} and the model-based Iterative SIC \cite{choi2000iterative} on 3 different channels, i.e., linear Gaussian (Eq.~\ref{eq:awgn}), non-linear quantized Gaussian (Eq.~\ref{eq:quant}), and non-linear Poisson (Eq.~\ref{eq:poisson}) channel. We focus on a scenario that the channel is highly dynamic with the signal-to-noise ratio (SNR) changing quickly and it would be unreasonable to train a neural network, e.g., \deepsic{}, from scratch for each SNR value. Therefore, we compare with a variant of \deepsic{} which is trained on a dataset of channel outputs from every SNR value in tested range $[0,14]$. For each SNR, we collect 5000 pairs of transmitted and received signals similarly to \cite{shlezinger2020deep} and call this variant \deepsic\textsf{(dynamic)}. We also include the results of the original \deepsic{}, that is trained and tested on each SNR value and denote this by \deepsic\textsf{(static)} for references to the situation that we can afford to retrain \deepsic{} every time.

Similarly to \cite{shlezinger2020deep}, we consider a $4 \times 4$ MIMO channel with $4$ transmitters and $4$ receive antennas, and binary phase shift keying (BPSK) constellation set. The channel matrix $\mathsf{H}$ for all three channel models follow spatial exponential decay, where the entry $H_{i,j} = e^{-|i-j|}, i \in \{ 1, \dots, R \}, j \in \{1, \dots, K\}$. We consider a small pilot signal of $10$ received sequences. For \method{}, the random sequence $\mathbf{r}$ is of length $4$ with each element drawn from normal distribution.

\subsection{Results}

The comparison results of \method{} with 2 versions of \deepsic{} and the iterative SIC methods on the three channels are plotted in Figure~\ref{fig:ser}. The results demonstrate improvements of \method{} over both \deepsic{} and the iterative SIC approaches consistently across all the signal-to-noise (SNR) values and all the considered channels. Notably, the gap between \method{} and the realistic variant of \deepsic, \deepsic\textsf{(dynamic)}, on the highly dynamic channel consideration, is significantly wide, proving the advantages of our \method{} framework on these highly dynamic channels. Even when compared with the \deepsic{}\textsf{(static)} that is fully retrained for each SNR value, \method{} still maintains an ample gap thanks to the GAN combination that leads to online training capability.

There is also a big gap between \deepsic\textsf{(dynamic)} and \deepsic\textsf{(static)} showing that adapting \deepsic{} method to the highly dynamic channels is not trivial. \deepsic\textsf{(dynamic)} is trained on an ensemble of received signals with all considered SNR values and performs poorly in the linear Gaussian and Poisson channels. Thus, \deepsic{} is only effective when being trained and tested on static channels. The model-based iterative SIC method works well on the linear Gaussian channel and the related quantized Gaussian channel while showing very deficient performance on the non-linear Poisson channel. This observation complies with the strong theoretical justification of iterative SIC on the linear Gaussian channel \cite{choi2000iterative} and no guarantees on the non-linear ones.

\section{Conclusion}

This work proposes to combine data-driven neural approach for symbol detection and the emerging adversarial neural networks (GANs) into an online training framework that is efficient on highly dynamic channels. We demonstrated improved performance of our framework compared to the latest neural networks and model-based approaches on various linear and non-linear channels.

\IEEEtriggeratref{30}
\bibliographystyle{IEEEtran}
\bibliography{relatedwork}

\end{document}